\documentclass[letterpaper, 10 pt, conference]{ieeeconf}  

\IEEEoverridecommandlockouts                              

\usepackage{graphics} 
\usepackage{epsfig} 
\usepackage{amsmath,amsfonts}
\usepackage[caption=false,font=normalsize,labelfont=sf,textfont=sf]{subfig}
\usepackage{textcomp}
\usepackage{stfloats}
\usepackage{url}
\usepackage{verbatim}
\usepackage{graphicx}
\usepackage{cite}
\usepackage{multirow}
\usepackage{xcolor}
\usepackage{bbding}
\usepackage{etoolbox}
\usepackage{cuted}
\usepackage{algorithmic}
\usepackage{caption}
\usepackage{hyperref}
\usepackage[dvipsnames]{xcolor}

\title{\LARGE \bf
BiGraspFormer: End-to-End Bimanual Grasp Transformer
}

\author{%
    Kangmin Kim$^{1}$,
    Seunghyeok Back$^{2}$,
    Geonhyup Lee$^{1}$,
    Sangbeom Lee$^{1}$,
    Sangjun Noh$^{1}$,
    Kyoobin Lee$^{1}$\textsuperscript{\dag}%
    \thanks{*This work was partly supported by Institute of Information $\&$ com-
            munications Technology Planning $\&$ Evaluation (IITP) grant funded by
            the Korea government (MSIT) (No.2019-0-01842, Artificial Intelligence
            Graduate School Program (GIST)) and Korea government(MSIT) (No. RS-
            2024-00459435, Core Technology Development of Open Simulator for SDx
            Intelligent Application Development)}
    \thanks{$^{1}$ Department of AI Convergence, Gwangju Institute of Science and Technology (GIST), Gwangju 61005, Republic of Korea.}%
    \thanks{$^{2}$ Department of AI Machinery, Korea Institute of Machinery \& Materials (KIMM), Daejeon 34103, Republic of Korea.}%
    \thanks{\textsuperscript{\dag}\,Corresponding author: Kyoobin Lee (\texttt{kyoobinlee@gist.ac.kr}).}%
}

\begin{document}

\maketitle
\thispagestyle{empty}
\pagestyle{empty}

\begin{abstract}


Bimanual grasping is essential for robots to handle large and complex objects. However, existing methods either focus solely on single-arm grasping or employ separate grasp generation and bimanual evaluation stages, leading to coordination problems including collision risks and unbalanced force distribution. To address these limitations, we propose BiGraspFormer, a unified end-to-end transformer framework that directly generates coordinated bimanual grasps from object point clouds. Our key idea is the Single-Guided Bimanual (SGB) strategy, which first generates diverse single grasp candidates using a transformer decoder, then leverages their learned features through specialized attention mechanisms to jointly predict bimanual poses and quality scores. This conditioning strategy reduces the complexity of the 12-DoF search space while ensuring coordinated bimanual manipulation. Comprehensive simulation experiments and real-world validation demonstrate that BiGraspFormer consistently outperforms existing methods while maintaining efficient inference speed ($<$0.05s), confirming the effectiveness of our framework. Code and supplementary materials are available at \href{https://sites.google.com/view/bigraspformer}{https://sites.google.com/bigraspformer}


\end{abstract}

\section{INTRODUCTION}

Bimanual grasping enables robots to manipulate large, heavy, or unwieldy objects beyond single-arm capabilities, making it essential for tasks such as lifting furniture, carrying long boards, or moving large boxes~\cite{gbagbe2024bi, billard2019trends}. However, most robotic grasping research has focused on single-arm systems, primarily on learning to detect 6-DoF grasp poses from point clouds~\cite{fang2020graspnet, liang2019pointnetgpd, sundermeyer2021contact, mahler2017dex, wu2020grasp, alliegro2022end}. While effective for single-arm tasks, these approaches cannot be directly extended to bimanual scenarios. First, bimanual grasping expands the action space to 12-DoF, doubling the computational complexity. Second, it introduces new challenges, including collision avoidance, balanced force/torque distribution, and dual-arm coordination for post-grasp manipulation.

\begin{figure}[!t]
    \centering
    \includegraphics[width=\columnwidth]{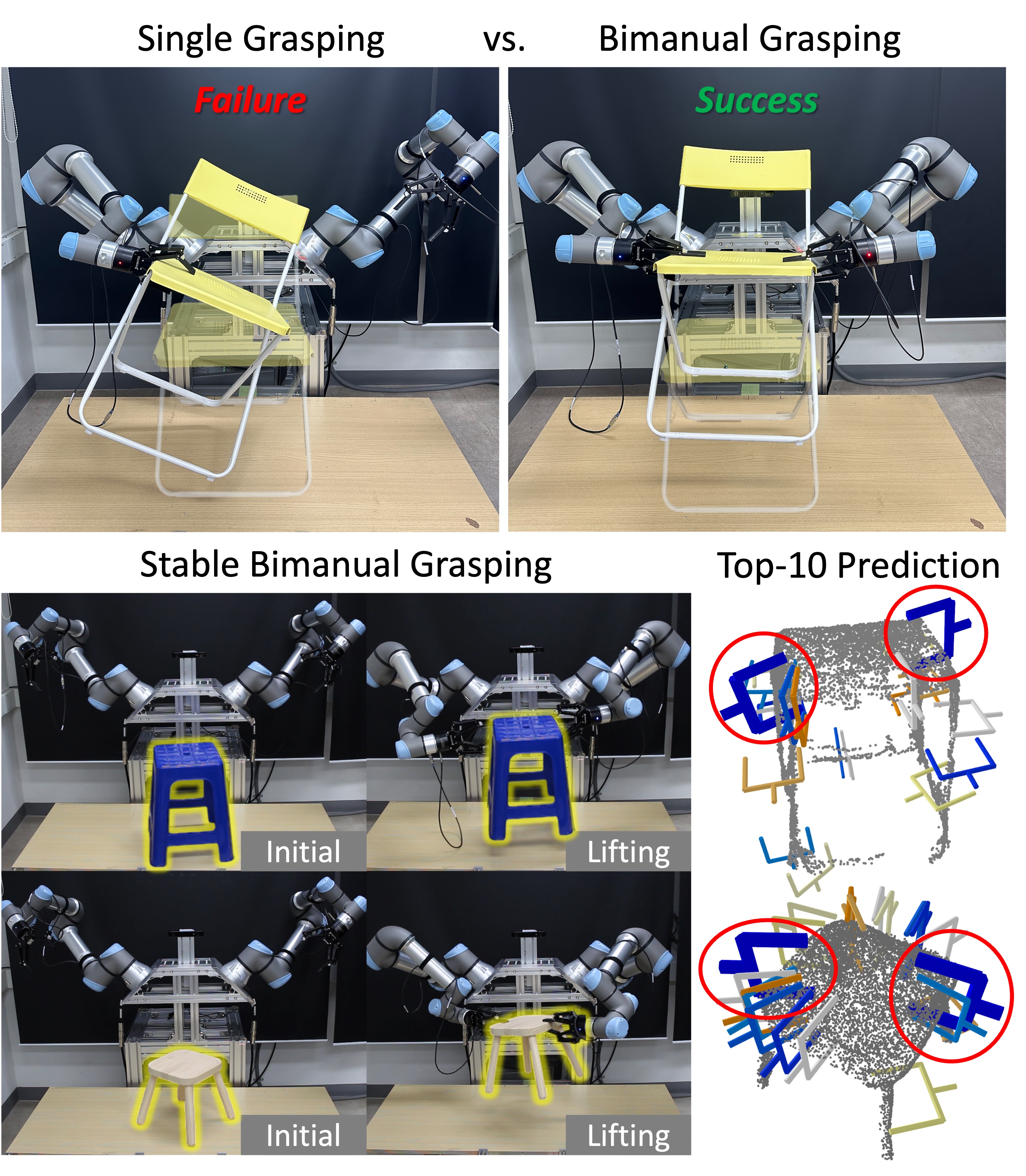}
    \caption{
    \textbf{BiGraspFormer for coordinated bimanual grasping.} (Top) Comparison between single and bimanual grasping for large objects. (Bottom) BiGraspFormer successfully grasps and lifts diverse objects in real-world environments, demonstrating stable grasps across various geometries. We visualize the point cloud with the top-10 grasp predictions; the highest-scoring pair is highlighted in thick blue.
    }
    \label{overview}
\end{figure}

For bimanual grasping, only a few methods have been proposed so far. The DA2 dataset~\cite{zhai20222} introduced the first benchmark by extending single-arm datasets~\cite{back2025graspclutter6d, eppner2021acronym, fang2020graspnet} with dual-arm-specific metrics such as force closure, dexterity, and torque balance~\cite{zhai20222, karim2025dg16m}. However, most existing approaches adopt modular architectures that separate grasp generation and evaluation. For example, Dual-PointNetGPD~\cite{zhai20222} evaluates the quality of grasp pairs from given candidates, requiring external single-arm grasp generators. Similarly, CGDF~\cite{singh2024constrained} directly generates bimanual grasps but lacks integrated quality prediction, instead relying on additional scoring modules or heuristic pairing strategies~\cite{borst2004grasp, ferrari1992planning}. As a result, current methods yield limited diversity, poor coordination, and high computation due to modular pipelines.

In this paper, we propose \textbf{BiGraspFormer}, the first unified end-to-end framework that directly generates coordinated bimanual grasps from object point clouds (Fig.~\ref{overview}). The key insight is that single-grasp features can effectively guide bimanual grasp generation, rather than treating dual-arm coordination as two independent problems. BiGraspFormer introduces a novel Single-Guided Bimanual (SGB) strategy: it first generates diverse single-arm grasp candidates, then leverages their learned features through specialized attention mechanisms to jointly predict bimanual poses and quality scores. This unified approach eliminates separate modules and explicitly models coordination between grasps, enabling stable and efficient dual-arm manipulation. Comprehensive experiments in both simulation and real-world environments demonstrate that BiGraspFormer achieves superior success, diversity, and speed compared to existing methods.

Our contributions are summarized as follows:

\begin{itemize}
\item We propose \textbf{BiGraspFormer}, the first unified end-to-end transformer for diverse, stable bimanual grasp generation.
\item We introduce the \textbf{Single-Guided Bimanual (SGB)} strategy, which leverages single-arm grasp features to guide bimanual generation, reducing computational complexity and enhancing dual-arm coordination.  
\item We achieve \textbf{state-of-the-art bimanual grasping performance} while maintaining fast inference suitable for real-world deployment.

\end{itemize}

\section{Related Works}

\subsection{Learning-based Single-arm Grasping}

Most existing methods have focused on single-arm grasping, evolving from discriminative approaches~\cite{liang2019pointnetgpd,mahler2017dex} that score grasp candidates to generative methods~\cite{mousavian20196, wu2020grasp, sundermeyer2021contact, alliegro2022end, wu2024economic} that directly synthesize grasps from visual inputs. While these single-arm approaches achieve strong performance on diverse objects, they face fundamental limitations for bimanual scenarios.  The action space expands from 6-DoF to 12-DoF, and coordinated dual-arm manipulation introduces critical requirements including collision avoidance, force balance, and post-grasp manipulability for task execution. Simply combining two independent single-arm solutions cannot address these coordination challenges. Our SGB strategy bridges this gap by leveraging single grasp features as guidance for coordinated bimanual generation, rather than simply pairing independent single-arm solutions.

\subsection{Model-based Bimanual Grasping}
Model-based bimanual grasping methods~\cite{tao2022optimal,vahrenkamp2011bimanual,cheraghpour2009multiple,caccavale2013grasp} have traditionally required explicit object knowledge such as 3D CAD models or predefined shape categories. These methods employ techniques including genetic algorithms for rod-shaped objects~\cite{tao2022optimal}, medial axis transforms for shape analysis\cite{vahrenkamp2011bimanual}, grasp matrix computations for stability evaluation~\cite{cheraghpour2009multiple,caccavale2013grasp}. 
While these approaches successfully generated stable grasps for known objects, they cannot handle unknown objects in unstructured environments due to their dependence on explicit geometric models. This fundamental limitation has motivated the development of learning-based approaches that can generalize to diverse unseen objects.

\subsection{Learning-based Bimanual Grasping}
Learning-based approaches have addressed model-based limitations through specialized datasets and neural methods. The DA2 dataset~\cite{zhai20222}
introduced dual-arm-specific metrics including force closure, dexterity, and torque balance, along with Dual-PointNetGPD for grasp quality evaluation. However, this approach only evaluates pre-generated candidates, requiring separate grasp generators for practical use. Generative methods have since emerged to address this limitation. Dual-Afford~\cite{zhao2022dualafford} learns collaborative affordance for dual-gripper manipulation, but requires explicit task labels and is trained on fixed task categories, limiting its generalization to arbitrary objects without task-specific supervision. CGDF~\cite{singh2024constrained} employs diffusion models to generate single grasps but requires additional pairing modules to form bimanual candidates, causing computational overhead and grasp pairs with potential arm collisions and unbalanced force distribution due to independent single-arm generation. BiGraspFormer addresses these limitations through unified end-to-end generation and evaluation, eliminating the need for separate modules while ensuring coordinated bimanual planning.




\begin{figure*}[!t]
\centering
    \includegraphics[width=\textwidth]{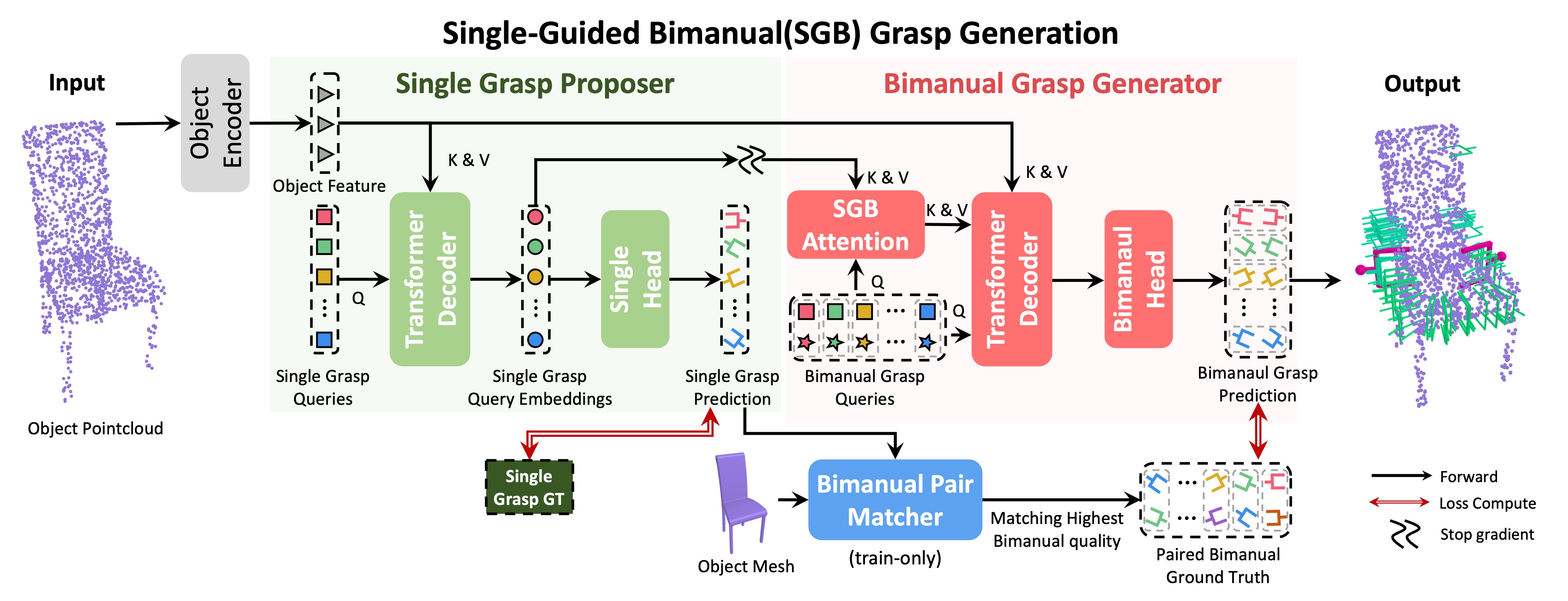}
\caption{
\textbf{Overview of our BiGraspFormer framework.}
An object encoder processes point cloud $P$ to extract geometric features. Single Grasp Proposer generates force-closure single grasps, Bimanual Pair Matcher matches them using bimanual quality metrics (force stability, torque balance, dexterity) to create ground truth, and Bimanual Grasp Generator employs SGB attention for final bimanual grasp generation.
}
\label{bigraspformer}
\end{figure*}

\section{Method}

\subsection{Motivation}
Our goal is to predict bimanual grasps $\mathcal{B} = \{\mathbf{g}^1, \mathbf{g}^2, q\}$ from an object point cloud $P$, where $\mathbf{g}^1$ and $\mathbf{g}^2$ represent the 6-DoF grasp poses of the two arms and $q$ denotes grasp quality. This task is challenging as it involves a 12-DoF action space, doubling the complexity of single-arm grasping. Each grasp needs to achieve force-closure stability while both arms coordinate to avoid collisions, maintain torque balance, and ensure overall stability.



To tackle this challenge, we introduce the \textbf{Single-Guided Bimanual (SGB)} grasp generation scheme, which decomposes the prediction of $\mathcal{B}$ into three structured stages. 
First, generate diverse single grasp candidates under basic stability constraints. Second, select feasible grasp pairs by discarding collisions and ensuring balanced force distribution. Finally, refine these pairs into stable bimanual grasps using learned features from both the object and single grasps. This formulation explicitly enforces both individual grasp quality and dual-arm coordination, decomposing the complex 12-DoF search space into a sequence of more tractable subproblems.

\subsection{BiGraspFormer}
\label{sec:bigraspformer}

BiGraspFormer employs the proposed SGB strategy as an end-to-end framework for bimanual grasp generation. As shown in Fig.~\ref{bigraspformer}, the network consists of four modules: 1) object encoder that extracts both local and global geometric features from the input point cloud, 2) Single Grasp Proposer (SGP) that generates diverse 6-DoF grasp candidates, 3) Bimanual Pair Matcher (BPM) that selects feasible grasp pairs based on collision checks and bimanual quality scores, and 4) Bimanual Grasp Generator (BGG) that refines these pairs into final 12-DoF bimanual grasps with scores.

\textbf{Object Encoder.} Bimanual grasping requires not only capturing fine-grained local geometry (e.g., contact points for force closure) but also modeling global object structure for coordinated bimanual grasps. We designed our object encoder to capture both local detail and global context within an integrated representation.
We use PointNet++~\cite{qi2017pointnet++} to extract local geometry. It uses Set Abstraction (SA) layers to get multi-scale features from the input point cloud $P \in \mathbb{R}^{N \times 3}$. This keeps the detailed shape information important for checking if single grasps are stable.
To complement these local features with global context, a transformer encoder is employed, consisting of self-attention layers. In this encoder, queries $q$, keys $k$, and values $v$ are derived from the local features from  PointNet++. This enables the network to capture relationships between distant regions and spatial relationships across the object. 
Specifically, the encoder consists of two SA layers from PointNet++ followed by six transformer encoder blocks. The final output is a set of object global features $\mathcal{F}'_g \in \mathbb{R}^{N' \times C}$ , which serve as the backbone representation for the subsequent stages of SGB grasp generation.


\textbf{Single Grasp Proposer.} 
Following the SGB scheme, the SGP generates diverse 6-DoF single-grasp candidates that satisfy force-closure constraints only. These candidates are subsequently paired and refined by the BPM and BGG modules into final bimanual grasps.
To this end, a DETR-style set prediction formulation~\cite{carion2020end} is employed, where a fixed set of learned object queries are decoded in parallel into output predictions without the need for hand-crafted anchors or post-processing. Specifically, $K'$ grasp queries $\{q^s_i\}_{i=1}^{K'}$ are randomly initialized learnable embeddings optimized end-to-end during training, and fed into a transformer decoder that attends to the encoded object features $\mathcal{F}'_g$. To enable query-based feature extraction, we employ cross-attention, where the grasp queries serve as $q$, while the encoded features $\mathcal{F}'_g$ serve as both keys $k$ and values $v$. This enables each learnable query to specialize in different object regions and capture distinct grasp poses. 
The decoder consists of six transformer decoder blocks, followed by a prediction head composed of three MLP layers. This module outputs a set of single grasp candidates $\hat{\mathcal{G}} = \{\hat{\mathbf{g}}_i\}_{i=1}^{K'}$, which provide the input for the subsequent pairing stage.


\textbf{Bimanual Pair Matcher.}
BPM bridges single grasp proposals and bimanual grasp generation by constructing reliable supervision pairs from the set $\hat{\mathcal{G}}$. Although the DA2 dataset~\cite{zhai20222} contains substantial grasp annotations (2,001 per object), the large object sizes make grasp coverage relatively sparse compared to single-arm datasets, making direct training insufficient for learning diverse bimanual coordination. BPM addresses this limitation by matching high-quality bimanual ground-truth pairs from the diverse single grasps proposed by the SGP. BPM operates in three steps: First, it evaluates all possible pairs from single grasps generated by SGP using the established bimanual quality metric from the DA2 dataset~\cite{zhai20222}, which combines force closure, dexterity, and torque balance. Second, collision checking eliminates infeasible pairs that would cause gripper-gripper or gripper-object collision using the object CAD model. Finally, for each anchor grasp, BPM selects the highest-scoring collision-free counterpart, forming reliable bimanual supervision pairs $\mathcal{B}^+={(\mathbf{g}^{1+}_i,\mathbf{g}^{2+}_i,q^+_i)}_{i=1}^{K'}$ as \textit{ground truth} for BGG. Note that BPM is used only during \textit{training} since it requires the object mesh. The process is efficiently parallelized on GPUs, introducing negligible computational overhead to the training pipeline.
Since BPM requires object CAD models for collision detection and ground truth quality evaluation, it cannot be deployed at inference time where only point cloud observations are available. Instead, during inference, the BGG module directly predicts both bimanual grasp poses and their quality scores in an end-to-end manner.



\textbf{Bimanual Grasp Generator.} 
As the final stage of the SGB framework, BGG generates 12-DoF bimanual grasps conditioned on single grasp features. It jointly refines them into complete bimanual poses with a quality that accounts for force closure, dexterity, and torque balance. 
BGG employs a DETR-style set prediction with learnable bimanual grasp queries $\{\mathbf{q}^b_i\}_{i=1}^{M'}$ processed through a transformer decoder equipped with the proposed \textbf{SGB attention layer}. In this layer, bimanual grasp queries serve as queries while single grasp features $\mathbf{F}_{sgp}$ from SGP serve as both keys and values, enabling bimanual generation to be conditioned on single grasp information:

\begin{equation}
\begin{split}
\text{SGB-Attention}(\mathbf{Q}^b, \mathbf{F}_{sgp}) \\ 
 = \text{softmax}&\left(\frac{\mathbf{Q}^b (\mathbf{F}_{sgp})^T}{\sqrt{d}}\right) \mathbf{F}_{sgp}
\end{split}
\end{equation}
where $\mathbf{Q}^b$ represents the bimanual grasp queries $\{\mathbf{q}^b_i\}_{i=1}^{M'}$ and $d$ is the embedding dimension. This allows each bimanual query to selectively attend to relevant single grasp information for coordinated bimanual grasp generation.

The SGB attention output $\mathcal{F}_{\mathrm{sgb}}$ and global object features $\mathcal{F}'_g$ both serve as key-value pairs for subsequent transformer decoder blocks. In each block, bimanual queries perform cross-attention to these features separately, fusing single-grasp evidence with global context. 
The decoder consists of six transformer blocks followed by a prediction head consisting of seven MLP layers, outputting bimanual candidates $\hat{\mathcal{B}} = \{(\hat{\mathbf{g}}^1_i, \hat{\mathbf{g}}^2_i, \hat{q_i})\}_{i=1}^{M'}$ where $\hat{\mathbf{g}}^1_i$ and $\hat{\mathbf{g}}^2_i$ are predicted 6-DoF grasp poses for both arms and $\hat{q}_i$ is the quality score.

\subsection{Implementation Details}
\textbf{Loss Functions.}  We utilize $\mathcal{L}_{dist}$ to regress grasp representation point $\mathbf{v}$, which is suggested in~\cite{sundermeyer2021contact}. The regression loss for single grasp:
\begin{equation}
\begin{split}
\mathcal{L}_{single}\left(\mathcal{G}_{i}, \hat{\mathcal{G}}_{j}\right) 
&= \mathcal{L}_{dist}\left(\mathbf{g}_{i}, \hat{\mathbf{g}}_{j}\right) \\
&= \left\| \mathbf{v}_{i} - \hat{\mathbf{v}}_j\right\|_2,
\end{split}
\end{equation}
For bimanual grasps, we incorporate an L1 loss as $\mathcal{L}_{quality}$ to regress the bimanual quality. The overall bimanual grasp loss is:
\begin{equation}
\begin{split}
\mathcal{L}_{bimanual}\left(\mathcal{B}_{i}, \hat{\mathcal{B}}_{j}\right)
 &= \mathcal{L}_{dist}\left(\mathbf{g}^{1}_{i}, \hat{\mathbf{g}}^{1}_{j}\right) 
 + \mathcal{L}_{dist}\left(\mathbf{g}^{2}_{i}, \hat{\mathbf{g}}^{2}_{j}\right) \\
 &+ \mathcal{L}_{quality}\left(q_{i}, \hat{q}_{j}\right)
\end{split}
\end{equation}

To enable training with these losses, we employ bipartite matching \cite{alliegro2022end} between predictions and ground truths. We use 
 the Hungarian method \cite{kuhn1955hungarian} to find optimal assignments, where the cost function for single grasps equals $\mathcal{L}_{single}$, while for bimanual grasps we exclude quality terms from $\mathcal{L}_{bimanual}$ during matching to focus on geometric alignment. The overall grasp loss for BiGraspFormer training is formulated as:
\begin{equation}
\mathcal{L}_{grasp}
= \mathcal{L}_{single}\left( \mathcal{G}_{i}, \hat{\mathcal{G}}_{\hat{\rho}_{s}i}\right)
+ \mathcal{L}_{bimanual}\left( \mathcal{B}_{i}, \hat{\mathcal{B}}_{\hat{\rho}_{bi}i}\right)
\end{equation}
where, $\hat{\rho}_{s}$ and $\hat{\rho}_{bi}$ are optimal assignments.

\textbf{Training Details.} 
The input point cloud consists of $N=2048$ points, with PointNet++~\cite{qi2017pointnet++} encoding them into $N'=512$ center points. The embedding dimension of global object features $C$ is set to 512. 
For training data preparation, we sample $K=128$ single grasps using the spatial grid-based method from~\cite{mousavian20196} to form uniformly distributed ground truth set $\mathcal{G}_{s}$. We set the number of single grasp queries $K'$, bimanual grasp queries $M'$, and ground truth bimanual grasps $M$ to 512 each.
The model is implemented in PyTorch 2.1 with CUDA 11.8 and trained on four NVIDIA Tesla A100 GPUs (40GB) using AdamW optimizer with learning rate $5 \times 10^{-4}$ and batch size 24 until validation loss convergence. Inference and real-world experiments use a single NVIDIA RTX 3090 GPU.

\begin{figure}[ht!]
    \centering
    \includegraphics[width=\columnwidth]{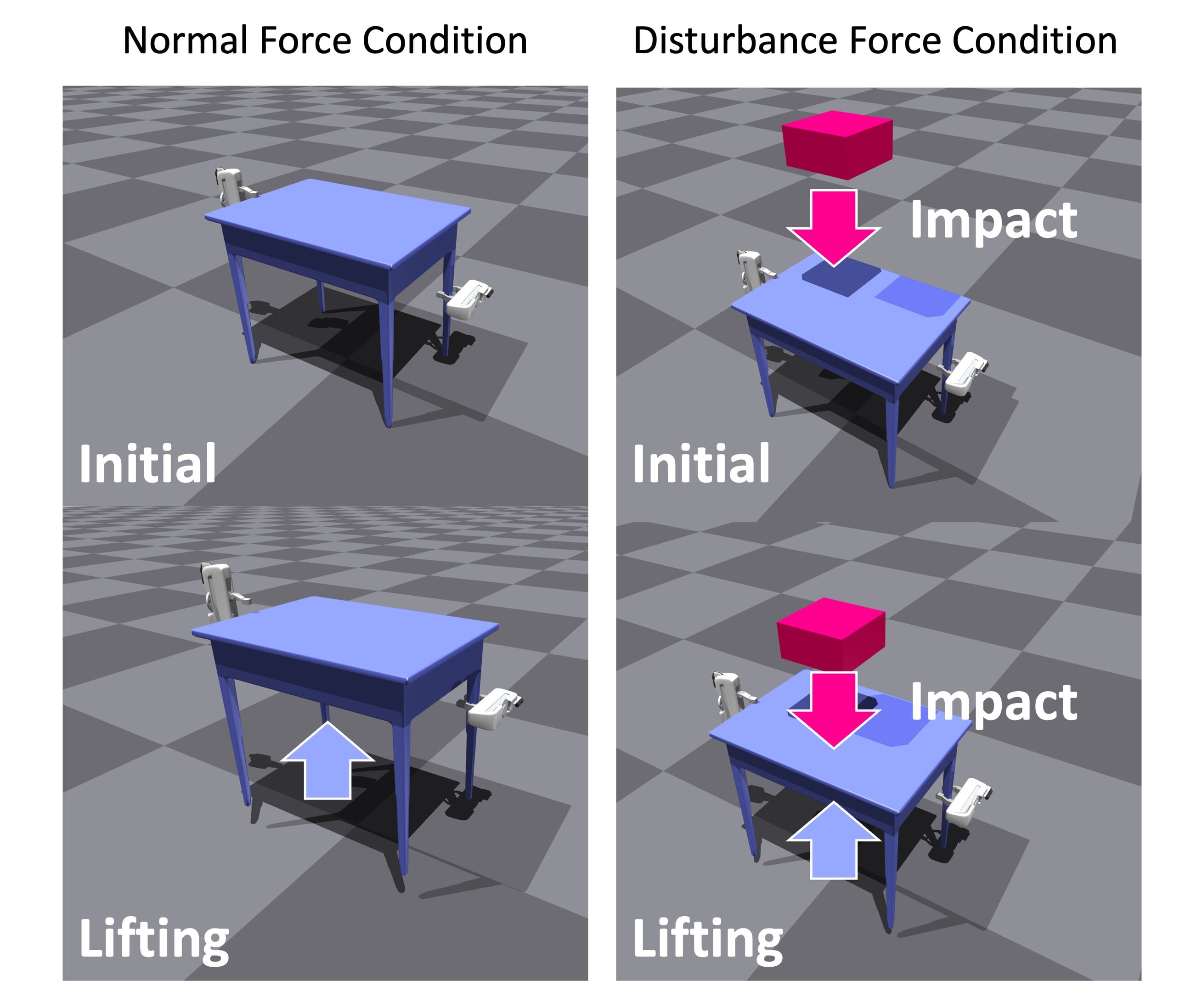}
    \caption{
    \textbf{Simulation experiment settings.} Left shows normal force condition where objects are grasped and lifted under gravity only. Right shows disturbance force condition where a weighted cube is dropped onto the object during lifting to apply additional external forces and test grasp robustness.
    }
    \label{tab:simulation env}
\end{figure}

\section{Experiments}

\subsection{Comparison with State-of-the-Art Methods}
\textbf{Datasets.}
We train and evaluate our BiGraspFormer using the DA2 dataset~\cite{zhai20222}, which contains 6,327 CAD models from ShapeNetSem~\cite{savva2015semantically}. We use the main files subset containing 3,417 objects, each with annotations of 2,001 bimanual grasps and corresponding quality scores.
Since no official data splits are provided, we divide the dataset into a training set with 2,823 objects, a validation set with 494 objects, and a test set of 100 objects. The test set consists of objects not included in the training or validation sets, used specifically for simulation-based evaluation. 
To train the SGP, we convert bimanual grasps into single grasps and remove duplicates. For the BGG, we generate ground truth bimanual grasps using BPM on the single grasps from the SGP, as described in Section ~\ref{sec:bigraspformer}.

\textbf{Metrics.}
For performance evaluation, we use two metrics: simulation-based grasp success rate and diversity. The success rate evaluation follows prior works~\cite{wu2020grasp, alliegro2022end, singh2024constrained, sundermeyer2021contact} under both normal force conditions and disturbance conditions where external forces are applied to test grasp robustness.
We utilize the Isaac Gym simulator~\cite{makoviychuk2021isaac} with two free-floating Franka Emika Panda grippers with extended 6-cm fingers.
Using grippers without arms ensures fair evaluation independent of robot reachability or approach direction~\cite{sundermeyer2021contact, mousavian20196, wu2020grasp, alliegro2022end, singh2024constrained}.
In the simulation, we place the grippers at the predicted bimanual grasp poses and close both simultaneously without gravity. After enabling gravity, both grippers lift the object upward. A grasp succeeds if all fingers maintain contact and the object remains firmly held.
For disturbance conditions, a 1kg weighted cube is dropped from 60cm height onto the center of the object during lifting to introduce additional external forces.

For diversity evaluation, we measure how well successful grasps cover the object surface. We approximate each gripper with a simplified geometric model~\cite{liang2019pointnetgpd} and position them at successful grasp poses from our predictions. For each object, we calculate diversity as the percentage of object point cloud points that lie within any gripper's geometric bounds, ensuring each point is counted only once regardless of how many grippers cover it. Higher diversity scores indicate that grasps are distributed across a larger portion of the object surface rather than concentrated in specific regions.
For these two metrics, we consider the predictions ranked in the top 1\%, 30\% and 50\% for grasp success rate, top 30\% and 50\% for diversity.

\begin{figure*}[htb!]
\centering
    \includegraphics[width=\textwidth]{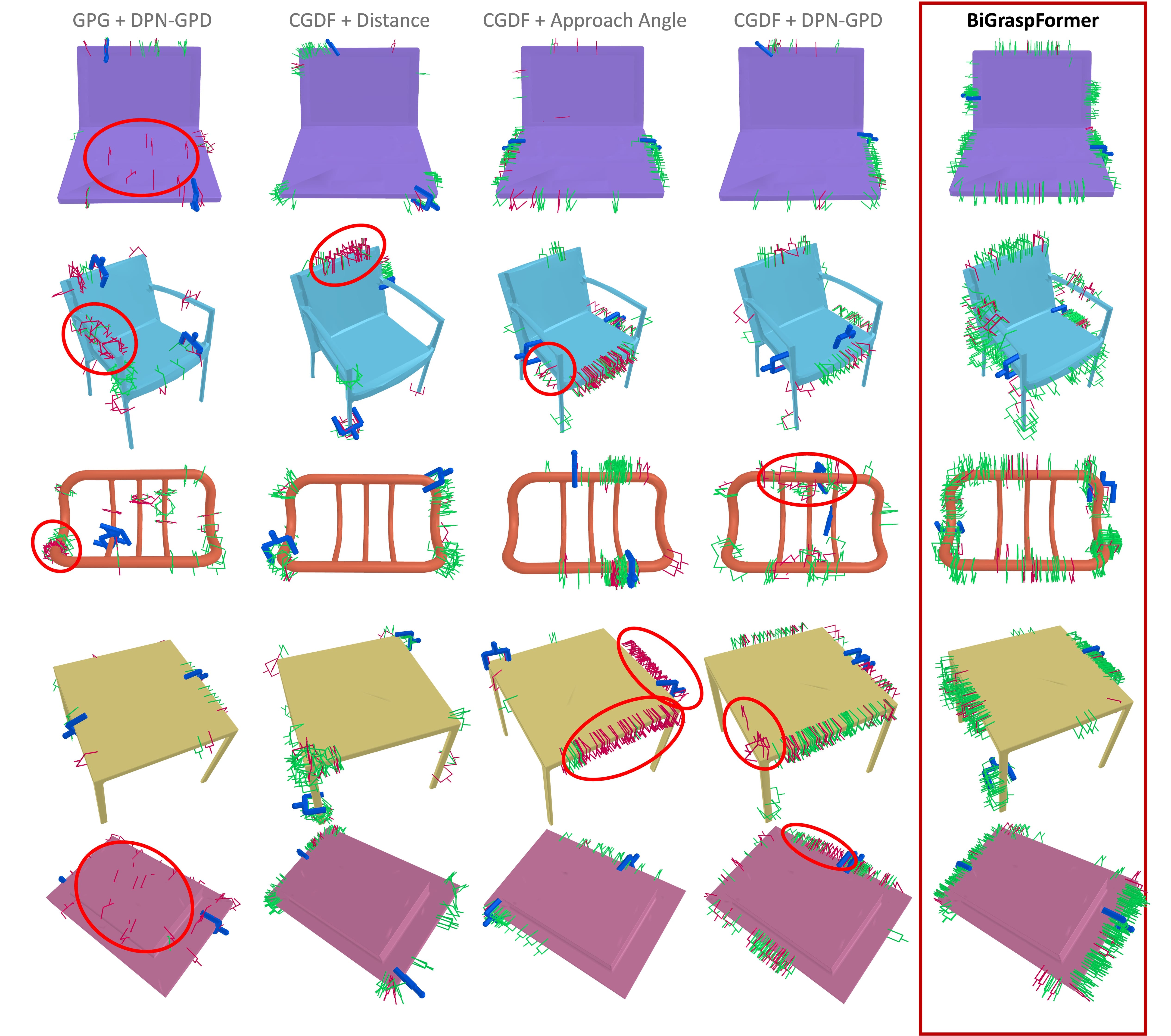}
    \caption{
    \textbf{Visualization of predicted bimanual grasp poses.} The top 100 bimanual grasp poses predicted by DPN-GPD, CGDF, and BiGraspFormer are shown for each object in the test set, based on simulation outcomes. The top-1 grasp pair is highlighted in blue. Green grasps represent successful grasps, while red grasps indicate failures due to instability, object collisions, or torque imbalance during grasping or lifting. Red circles highlight notable failure cases of baseline methods.
    }
\label{tab:qualitative result}
\end{figure*}

\textbf{Baselines.}
We used the following baselines:
\begin{itemize}
\item \textbf{GPG + DPN-GPD}: GPG~\cite{ten2018using} for single grasp generation with DPN-GPD~\cite{zhai20222} for quality-based bimanual pairing.

\item \textbf{CGDF + Distance}: Unconstrained CGDF~\cite{singh2024constrained} for single grasp generation with distance-based pairing that selects two grasps located farthest apart on the object~\cite{borst2004grasp, ferrari1992planning}.

\item \textbf{CGDF + Approach Angle}: Unconstrained CGDF for single grasp generation with approach angle-based pairing, selecting grasps with opposing approach directions similar to antipodal selection~\cite{eppner2019billion}.

\item \textbf{CGDF + DPN-GPD}: Unconstrained CGDF for single grasp generation with DPN-GPD for quality-based bimanual pairing.
\end{itemize}
For fair comparison, we use 256 single grasp candidates for all methods to balance computational efficiency with grasp diversity. DPN-GPD is trained to score grasp quality for bimanual scenarios, while unconstrained CGDF generates grasps across the entire object surface without restricting to specific regions. We select the top 512 bimanual pairs based on their predicted quality scores for evaluation.

\renewcommand{\arraystretch}{1.3} 
\begin{table}[]
\caption{Simulation based grasp success and diversity performance of BiGraspFormer and state-of-the-art method on disturbance force condition}
\label{tab:main table2}
\resizebox{\columnwidth}{!}{%
\begin{tabular}{l|rrr|rr}
\hline
\multicolumn{1}{c|}{\multirow{2}{*}{\normalsize Method}} & \multicolumn{3}{c|}{\normalsize \normalsize Success Rate} & \multicolumn{2}{c}{\normalsize Diversity} \\ \cline{2-6} 
\multicolumn{1}{c|}{} & \multicolumn{1}{c|}{\normalsize Top 1\%} & \multicolumn{1}{c|}{\normalsize Top 30\%} & \multicolumn{1}{c|}{\normalsize Top 50\%} & \multicolumn{1}{c|}{\normalsize Top 30\%} & \multicolumn{1}{c}{\normalsize Top 50\%} \\ \hline
GPG~\cite{ten2018using}+DPN-GPD~\cite{zhai20222} & \multicolumn{1}{r|}{\normalsize 20.96} & \multicolumn{1}{r|}{\normalsize 19.72} & \normalsize 19.97 & \multicolumn{1}{r|}{\normalsize 12.20} & \normalsize 14.91 \\
CGDF~\cite{singh2024constrained}+Distance~\cite{ferrari1992planning} & \multicolumn{1}{r|}{\normalsize 23.98} & \multicolumn{1}{r|}{\normalsize 24.02} & \normalsize 23.42 & \multicolumn{1}{r|}{\normalsize 8.28} & \normalsize 9.12 \\
CGDF~\cite{singh2024constrained}+Approach Angle~\cite{eppner2019billion} & \multicolumn{1}{r|}{\normalsize 34.70} & \multicolumn{1}{r|}{\normalsize31.34} & \normalsize31.77 & \multicolumn{1}{r|}{\normalsize15.51} & \normalsize18.17 \\
CGDF~\cite{singh2024constrained}+DPN-GPD~\cite{zhai20222} & \multicolumn{1}{r|}{\normalsize36.46} & \multicolumn{1}{r|}{\normalsize32.14} & \normalsize32.86 & \multicolumn{1}{r|}{\normalsize14.39} & \normalsize17.95 \\ \hline
\textbf{BiGraspFormer(Ours)} & \multicolumn{1}{r|}{\normalsize\textbf{59.72}} & \multicolumn{1}{r|}{\normalsize\textbf{57.16}} & \normalsize\textbf{55.66} & \multicolumn{1}{r|}{\normalsize\textbf{29.40}} & \normalsize\textbf{37.99} \\ \hline
\end{tabular}%
}
\end{table}

\begin{table}[]
\caption{Simulation based grasp success and diversity performance of BiGraspFormer and state-of-the-art method on normal force condition}
\label{tab:main table}
\resizebox{\columnwidth}{!}{%
\begin{tabular}{l|rrr|rr|r}
\hline
\multicolumn{1}{c|}{\multirow{2}{*}{\normalsize Method}} & \multicolumn{3}{c|}{\normalsize Success Rate} & \multicolumn{2}{c|}{\normalsize Diversity} & \multicolumn{1}{c}{\multirow{2}{*}{\begin{tabular}[c]{@{}c@{}} \normalsize Time\\ \normalsize (s)\end{tabular}}} \\ \cline{2-6}
\multicolumn{1}{c|}{} & \multicolumn{1}{c|}{\scriptsize \normalsize Top 1\%} & \multicolumn{1}{c|}{\scriptsize \normalsize Top 30\%} & \multicolumn{1}{c|}{\scriptsize \normalsize Top 50\%} & \multicolumn{1}{c|}{\scriptsize \normalsize Top 30\%} & \scriptsize \normalsize Top 50\% & \multicolumn{1}{c}{} \\ \hline
GPG~\cite{ten2018using}+DPN-GPD~\cite{zhai20222} & \multicolumn{1}{r|}{\normalsize53.05} & \multicolumn{1}{r|}{\normalsize53.32} & \normalsize53.37 & \multicolumn{1}{r|}{\normalsize17.23} & \normalsize19.65 & \normalsize10.99 \\
CGDF~\cite{singh2024constrained}+Distance~\cite{ferrari1992planning} & \multicolumn{1}{r|}{\normalsize76.09} & \multicolumn{1}{r|}{\normalsize78.01} & \normalsize77.75 & \multicolumn{1}{r|}{\normalsize12.66} & \normalsize13.18 & \normalsize4.39 \\
CGDF~\cite{singh2024constrained}+Approach Angle~\cite{eppner2019billion} & \multicolumn{1}{r|}{\normalsize71.38} & \multicolumn{1}{r|}{\normalsize71.14} & \normalsize71.04 & \multicolumn{1}{r|}{\normalsize23.29} & \normalsize26.13 & \normalsize4.82 \\
CGDF~\cite{singh2024constrained}+DPN-GPD~\cite{zhai20222} & \multicolumn{1}{r|}{\normalsize71.61} & \multicolumn{1}{r|}{\normalsize71.79} &\normalsize 72.67 & \multicolumn{1}{r|}{\normalsize20.74} & \normalsize24.26 & \normalsize18.18 \\ \hline
\textbf{BiGraspFormer(Ours)} & \multicolumn{1}{r|}{\normalsize\textbf{89.67}} & \multicolumn{1}{r|}{\normalsize\textbf{84.86}} &\normalsize \textbf{83.65} & \multicolumn{1}{r|}{\normalsize\textbf{36.99}} & \normalsize\textbf{46.47} & \normalsize\textbf{0.04} \\ \hline
\end{tabular}%
}
\end{table}

\textbf{Result}
Table~\ref{tab:main table2} and Table~\ref{tab:main table} compare grasp success rates and diversity with state-of-the-art methods in simulation. BiGraspFormer consistently outperforms all baselines under both normal and disturbance conditions. In particular, BiGraspFormer achieves superior top 1\% success rates across all conditions. 
Our method also demonstrates superior diversity compared to baselines, generating diverse yet stable bimanual grasp poses. Under external disturbances (Table~\ref{tab:main table2}), BiGraspFormer consistently surpasses baselines in both success and diversity, confirming robustness to disturbances.

Table~\ref{tab:main table} also reports computational efficiency measured on an RTX 3090 and Intel i7-12700K. For fair comparison, we measure total inference time including model forward pass for all methods, plus pairing stage time for baseline approaches that require separate pairing modules. Times are averaged across multiple runs. BiGraspFormer achieves inference times under 0.05 seconds, demonstrating efficient grasp generation. Fig.~\ref{tab:qualitative result} shows qualitative comparisons with state-of-the-art methods. BiGraspFormer generates more diverse and stable bimanual grasp candidates across various object geometries.

\begin{table}[h!]
\caption{Ablation study of the modules in the SGB framework on normal force condition}
\label{tab:ablation_table}
\resizebox{\columnwidth}{!}{%
\begin{tabular}{|cc|rrr|rr|}
\hline
\multirow{2}{*}{BPM} & \multirow{2}{*}{\begin{tabular}[c]{@{}c@{}}SGB\\ attention\end{tabular}} & \multicolumn{3}{c|}{Success Rate} & \multicolumn{2}{c|}{Diversity} \\ \cline{3-7} 
 &  & \multicolumn{1}{c|}{Top 1\%} & \multicolumn{1}{c|}{Top 30\%} & \multicolumn{1}{c|}{Top 50\%} & \multicolumn{1}{c|}{Top 30\%} & \multicolumn{1}{l|}{Top 50\%} \\ \hline
\textcolor{red}{\XSolidBrush} & \textcolor{red}{\XSolidBrush} & \multicolumn{1}{r|}{65.64} & \multicolumn{1}{r|}{60.15} & 60.37 & \multicolumn{1}{r|}{32.69} & 38.74 \\
\textcolor{ForestGreen}{\large \textbf{\checkmark}} & \textcolor{red}{\XSolidBrush} & \multicolumn{1}{r|}{80.61} & \multicolumn{1}{r|}{77.40} & 76.30 & \multicolumn{1}{r|}{36.92} & 46.18 \\
\textcolor{ForestGreen}{\large \textbf{\checkmark}} & \textcolor{ForestGreen}{\large \textbf{\checkmark}} & \multicolumn{1}{r|}{\textbf{89.67}} & \multicolumn{1}{r|}{\textbf{84.86}} & \textbf{83.65} & \multicolumn{1}{r|}{\textbf{36.99}} & \textbf{46.47} \\ \hline
\end{tabular}%
}
\end{table}

\begin{table}[h!]
\caption{Analysis of different numbers of bimanual grasp queries on normal force condition}
\label{tab:ablation_table2}
\resizebox{\columnwidth}{!}{%
\begin{tabular}{l|rrr|rr}
\hline
\multicolumn{1}{c|}{\multirow{2}{*}{\begin{tabular}[c]{@{}c@{}}Number of \\ Queries\end{tabular}}} & \multicolumn{3}{c|}{Success Rate} & \multicolumn{2}{c}{Diversity} \\ \cline{2-6} 
\multicolumn{1}{c|}{} & \multicolumn{1}{c|}{Top 1\%} & \multicolumn{1}{c|}{Top 30\%} & \multicolumn{1}{c|}{Top 50\%} & \multicolumn{1}{c|}{Top 30\%} & \multicolumn{1}{l}{Top 50\%} \\ \hline
128 & \multicolumn{1}{r|}{76.40} & \multicolumn{1}{r|}{70.28} & 68.78 & \multicolumn{1}{r|}{18.10} & 24.06 \\
256 & \multicolumn{1}{r|}{82.50} & \multicolumn{1}{r|}{79.10} & 78.37 & \multicolumn{1}{r|}{27.39} & 35.49 \\ \hline
\textbf{512(Ours)} & \multicolumn{1}{r|}{\textbf{89.67}} & \multicolumn{1}{r|}{\textbf{84.86}} & \textbf{83.65} & \multicolumn{1}{r|}{\textbf{36.99}} & \textbf{46.47} \\ \hline
\end{tabular}%
}
\end{table}

\subsection{Ablation Study}
\textbf{Effect of Single-Guided Bimanual Grasp Generation.}
We conduct ablation studies to verify the effectiveness of the proposed SGB framework by comparing BiGraspFormer with variants lacking the BPM and SGB attention layer. 
As a baseline, we use a model that directly generates bimanual grasps. This baseline consists of only the object encoder and BGG module without SGB attention layer, trained on the original DA2 dataset~\cite{zhai20222} instead of BPM-generated pairs.
Table~\ref{tab:ablation_table} shows that directly generating bimanual grasps fails to achieve state-of-the-art performance. Without the SGB attention layer (using only BPM), the model achieves competitive top-1\% success rates but degrades at top-30\% and top-50\%. In contrast, BiGraspFormer with the complete SGB framework outperforms all variants across all metrics, demonstrating the effectiveness of our SGB grasp generation scheme for producing stable bimanual grasp candidates. 

\textbf{Effect of number of Bimanual Grasp Queries}
To investigate the impact of query quantity, we compare BiGraspFormer models with varying numbers of learnable bimanual grasp queries (128, 256, and 512).
Table~\ref{tab:ablation_table2} shows that increasing the number of queries consistently improves both metrics, indicating that more queries better capture the complexity of the bimanual grasp space. With only 128 queries, the model performs worse than state-of-the-art baselines, suggesting that insufficient queries cannot adequately represent the large bimanual grasp space.
Notably, BiGraspFormer achieves state-of-the-art performance with just 256 queries. Even with half the candidates, it demonstrates higher diversity than other methods generating 512 bimanual pairs. This validates our SGB framework's effectiveness in decomposing the complex 12-DoF search space into simpler subproblems, enabling the model to learn more efficiently and generate more diverse grasps with fewer queries.



\begin{figure*}[ht!]
    \centering
    \includegraphics[width=\textwidth]{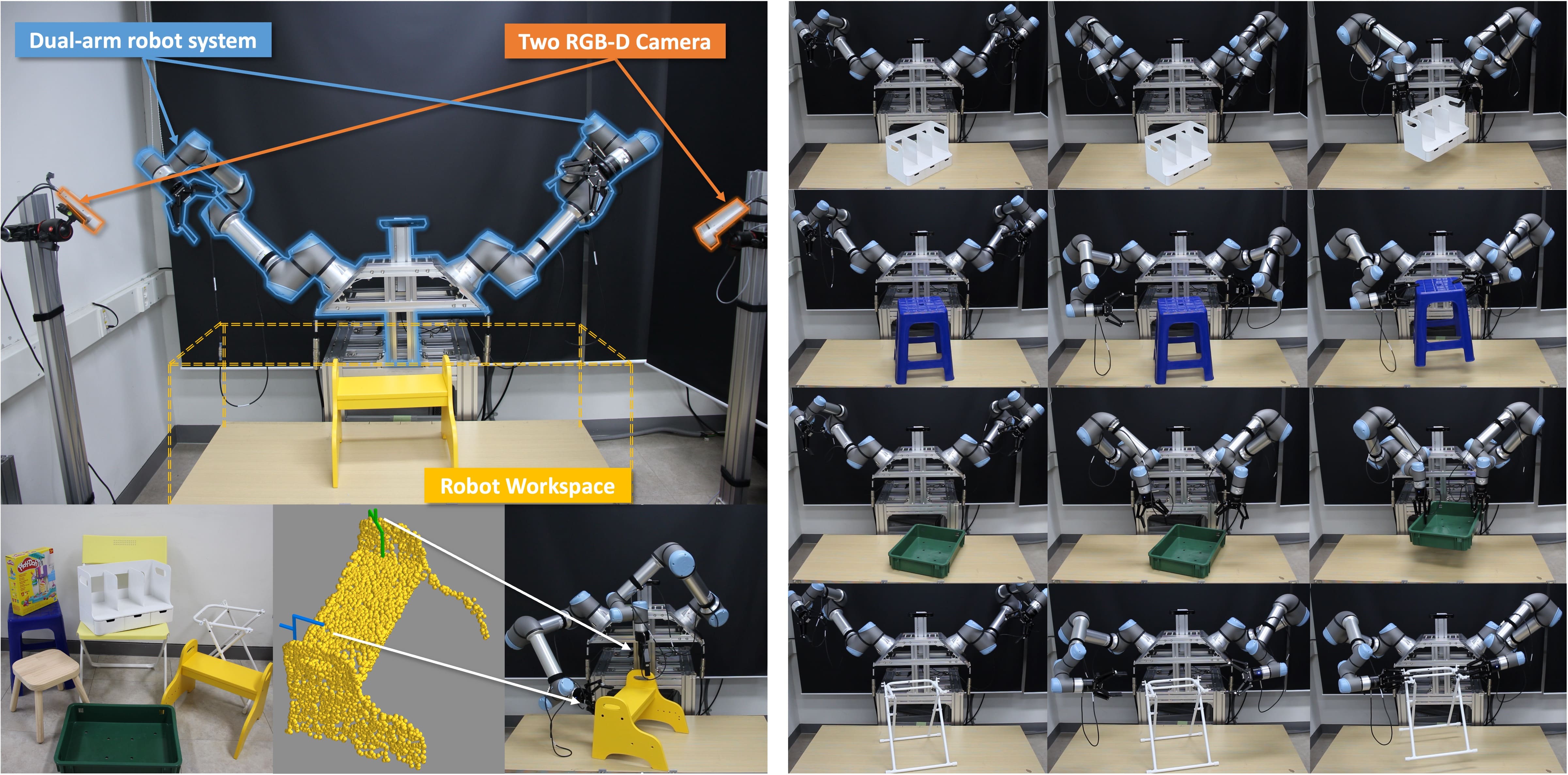}
   \caption{\textbf{Real-World Experimental Setup.} (Left) Dual-arm robotic system with two UR5e arms and two Azure Kinect RGB-D cameras, along with test objects and visualization of feasible bimanual grasps overlaid on object point clouds. (Right) Collision-free trajectory execution of selected grasp poses for grasping and lifting diverse objects.}
    \label{real setting}
\end{figure*}


\begin{table}[h!]
\renewcommand{\arraystretch}{1.2}
\centering
\caption{Grasp performance in the real-world}
\label{real_world_exp}
\resizebox{\columnwidth}{!}{%
\begin{tabular}{cccccccc}
\hline
\begin{tabular}[c]{@{}c@{}}Yellow\\ chair\end{tabular} & \begin{tabular}[c]{@{}c@{}}Wood\\ stool\end{tabular} & \begin{tabular}[c]{@{}c@{}}Green\\ bin\end{tabular} & \begin{tabular}[c]{@{}c@{}}White\\ shelf\end{tabular} & \begin{tabular}[c]{@{}c@{}}Toy\\ box\end{tabular} & \begin{tabular}[c]{@{}c@{}}Blue\\ chair\end{tabular} & \begin{tabular}[c]{@{}c@{}}White\\ frame\end{tabular} & \begin{tabular}[c]{@{}c@{}}Yellow\\ stair\end{tabular} \\ \hline
\multicolumn{1}{r}{8/10} & \multicolumn{1}{r}{10/10} & \multicolumn{1}{r}{10/10} & \multicolumn{1}{r}{8/10} & \multicolumn{1}{r}{10/10} & \multicolumn{1}{r}{9/10} & \multicolumn{1}{r}{9/10} & \multicolumn{1}{r}{7/10} \\ \hline
\end{tabular}%
}
\end{table}

\subsection{Real Robot Experiments}
To validate that BiGraspFormer generalizes beyond simulation, we conducted real-robot experiments evaluating grasp success on large, complex-shaped objects placed in diverse poses.
Our setup consisted of two UR5e robotic arms, each equipped with a Robotiq 2F-140 gripper, and two Azure Kinect cameras mounted on opposite sides of the workspace (Fig.~\ref{real setting}). 
Point clouds captured from both cameras were first cropped to the workspace region using RANSAC-based plane fitting~\cite{fischler1981random}, then transformed into the robot coordinate frame and registered into a unified point cloud via ICP~\cite{besl1992method}, which was provided to the model as input. In each trial, the target object was placed on a table in a different orientation and position. We employed curobo~\cite{sundaralingam2023curobo} for collision-free motion planning, using the real object point cloud. Grasp candidates predicted by BiGraspFormer were ranked by their quality scores, and the predicted grasps were evaluated in descending order of quality score. For each candidate, motion planning was attempted, and if a collision-free trajectory was found, the corresponding grasp was executed. The robot then lifted the object upward and placed it back down; the attempt was deemed successful if the object remained securely held throughout the lift.

We conducted 10 trials per object, each with a distinct object pose, across a total of eight test objects (Fig.~\ref{real setting}). As summarized in Table~\ref{real_world_exp}, BiGraspFormer successfully grasped and lifted most objects. For simpler large objects such as the toy box and green bin, all trials succeeded. For more challenging shapes such as the yellow chair, white frame, and blue chair, the model still achieved high success rates. 
Failures mainly stemmed from unstable torque distribution, occasionally leading to slippage during lifting.
The yellow stair was the most difficult object due to its heavy weight, where even minor imbalances caused grasp failures.
Despite these challenges, BiGraspFormer consistently demonstrated strong performance across all tested objects, confirming its ability to generate reliable bimanual grasps for large and complex objects in real-world environments.



\section{Conclusions}
We presented BiGraspFormer, the first unified end-to-end framework that directly generates coordinated bimanual grasps from object point clouds using our novel Single-Guided Bimanual (SGB) strategy. The key insight of SGB is to first generate diverse single grasp candidates then leverage their learned features through specialized attention mechanisms to jointly predict bimanual poses and quality scores, effectively reducing the complexity of the 12-DoF search space while ensuring coordinated dual-arm manipulation. This unified approach eliminates the need for separate grasp generation and evaluation modules, enabling explicit modeling of coordination between grasps for stable bimanual manipulation. Extensive simulation and real-world experiments demonstrate that BiGraspFormer consistently outperforms existing methods in both success rate and diversity under normal and disturbance conditions while maintaining efficient inference speed suitable for real-time deployment. Despite these promising results, our approach has several limitations. BiGraspFormer currently relies on CAD models during training for collision checking and ground truth generation via BPM, which may limit applicability in settings without high-quality object meshes. Additionally, the sim-to-real gap and motion planning failures remain challenges, as grasp poses that succeed in simulation may occasionally fail due to real-world sensor noise or kinematic constraints. Future work will focus on removing the CAD model dependency by incorporating learned collision proxies, as well as generating action-policy-aware bimanual grasps integrated with downstream manipulation policies to enable comprehensive bimanual manipulation tasks.









\bibliographystyle{IEEEtran}
\bibliography{references}

\end{document}